\documentclass[journal]{IEEEtran}
\usepackage{amsmath,amssymb,amsfonts}
\usepackage{algorithmic}
\usepackage{algorithm}
\usepackage{array}
\usepackage[caption=false,font=normalsize,labelfont=sf,textfont=sf]{subfig}
\usepackage{textcomp}
\usepackage{stfloats}
\usepackage{url}
\usepackage{verbatim}
\usepackage{graphicx}
\usepackage{cite}
\usepackage{multirow}
\usepackage{colortbl}
\usepackage[hidelinks]{hyperref}

\definecolor{red}{RGB}{255,0,0}
\definecolor{green}{RGB}{0,255,0}
\definecolor{blue}{RGB}{0,0,255}
\definecolor{orange}{RGB}{255,165,0}
\definecolor{title_gray}{gray}{.9}

\newcommand{\thickhline}{\noalign{\hrule height 1pt}}

\begin{document}

\title{
\begin{small}
Copyright © 2026 IEEE. Personal use of this material is permitted. However, permission to use this material for any other purposes must be obtained from the IEEE by sending an email to pubs-permissions@ieee.org. \\
\end{small}
Unified Modeling of Lane and Lane Topology for Driving Scene Reasoning}

\author{Han Li, Yulu Gao, Si Liu, Yuhang Wang, Bo Liu, Beipeng Mu
\thanks{
Han Li is with the School of Artificial Intelligence, Beihang University, Beijing, China, and also with Zhongguancun Academy, Beijing, China.  
E-mail: lihan0620@buaa.edu.cn.
}
\thanks{
Yulu Gao is with the Hangzhou International Innovation Institute, Beihang University, Hangzhou, China.
E-mail: gyl97@buaa.edu.cn.
}
\thanks{
Si Liu is with the School of Artificial Intelligence, Beihang University, Beijing, China.
E-mail: liusi@buaa.edu.cn.
}
\thanks{Yuhang Wang, Bo Liu, and Beipeng Mu are with Meituan, Beijing, China.
E-mails: \{wangyuhang11, mubeipeng\}@meituan.com, boliu1995@163.com.
}
\thanks{The first two authors contributed equally to this work, and the corresponding author is Si Liu.}
\thanks{Digital Object Identifier 10.1109/TCSVT.2026.3690152}
}

\markboth{IEEE TRANSACTIONS ON CIRCUITS AND SYSTEMS FOR VIDEO TECHNOLOGY} %
{H.~Li \MakeLowercase{\textit{et al.}}: Unified Modeling of Lane and Lane Topology for Driving Scene Reasoning}

\maketitle

\begin{abstract}
Autonomous vehicles need to perceive not only physical elements in the driving scene, such as lane lines and traffic lights, but also logical elements like lane centerlines and their topology.
Existing lane topology reasoning methods typically follow a reasoning-by-detection paradigm, where lane topological relationships are primarily derived from lane detection results.
In this paper, we propose an innovative method called Unified Modeling of Lane and Lane Topology~(UniTopo), which represents the topological relationships between lanes as connected lanes, encompassing predecessor lanes, successor lanes, and their interconnections.
This unified representation of lanes and lane topology allows us to simultaneously obtain both the positions and topological information of lanes within a shared perception pipeline, establishing a new paradigm for directly perceiving lane topology from original image features.
We validate our method on the driving scene reasoning benchmark OpenLane-V2, which consists of two subsets, built based on Argoverse2 and nuScenes, respectively.
Our method achieves $\text{TOP}_{ll}$ of 30.1\% and 31.8\% on the two subsets, significantly surpassing the existing state-of-the-art method $\text{T}^2\text{SG}$ by 6.0\% and 8.6\%.
The code is available at \url{https://github.com/homothetic/UniTopo}.
\end{abstract}


\section{Introduction}
Lane lines, road boundaries, traffic lights, and road markers are essential physical components in the driving environment, playing a pivotal role in ensuring the safe operation of vehicles~\cite{nuScenes, Argoverse2, waymo}. 
With the development of autonomous driving technology, modern advanced driver assistance systems not only need to accurately perceive these physical elements, but also achieve a higher level of scene comprehension.
This encompasses predicting logical elements such as lane centerlines and their topology, as well as the relationships between lane centerlines and traffic elements.
The logical elements within the driving scene enable autonomous vehicles to comprehend complex road structures, identify drivable areas, and support downstream tasks such as planning and control~\cite{UniAD, VAD}.

Recently, with the release of the OpenLane-V2~\cite{OpenLane-V2} benchmark, a growing number of studies have focused on achieving both perception and reasoning tasks within end-to-end networks.
Existing methods~\cite{TopoNet, TopoMLP, Topo2D, TopoLogic, roadpainter, TSTGT, TopoFormer, SMERF, LaneSegNet} can be categorized into two mainstream frameworks.
The first framework~\cite{TopoMLP, Topo2D} uses a separate network to detect lane centerlines, and then employs Multi-Layer Perceptrons~(MLPs) to predict lane topological relationships.
The second framework~\cite{TopoNet, TopoLogic} introduces topological information into the detection pipeline. For instance, TopoNet~\cite{TopoNet} employs Graph Convolutional Network~(GCN)~\cite{GCN} within the lane decoder to model lane topological relationships.

\begin{figure}[t]
  \centering
  \includegraphics[width=1.0\linewidth]{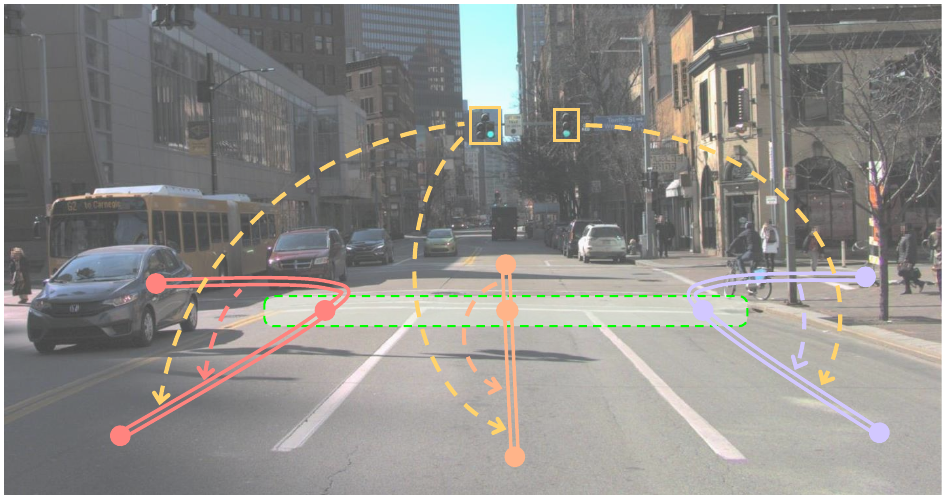}
  \caption{
  \textbf{Motivation of UniTopo.} 
  Unlike the topological relationships between lanes and traffic elements, the connections among piecewise lanes can be observed in images by identifying where the lanes connect.
  These areas are highlighted by the \textcolor{green}{green} rectangle (e.g., regions containing junction points near stop lines), and all topological relations are explicitly visualized using dashed arrows. 
  }
  \label{fig:motivation}
\end{figure}

Despite their differences, these methods~\cite{TopoNet, TopoMLP, Topo2D, TopoLogic} follow a reasoning-by-detection paradigm, where detecting lanes is considered a prerequisite for lane topology reasoning.
However, as shown in Figure~\ref{fig:motivation}, unlike the topological relationships between lanes and traffic elements, the connections among piecewise lanes are clear and observable.
It is more intuitive to perceive lane-to-lane topology directly from the original image features, rather than first detecting all lanes using a detection model and then inferring their topological relationships by pairing these lanes based on measures such as feature similarity and geometric distances.

\begin{figure}[t]
  \centering
  \includegraphics[width=\linewidth]{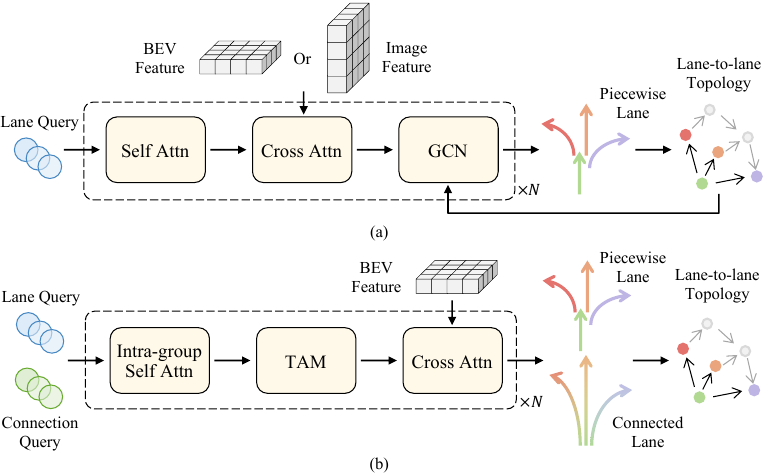}  
  \caption{(a) Previous methods follow a reasoning-by-detection paradigm, where lane detection is regarded as a prerequisite for lane topology reasoning. (b) Our method captures the connectivity relationships between piecewise lanes by defining connection queries that interact with BEV features within a shared lane detection pipeline.}
  \label{fig:compare}
\end{figure}

Inspired by this insight, we propose a method called \textbf{Uni}fied Modeling of Lane and Lane \textbf{Topo}logy~(UniTopo), which depicts lane-to-lane topology as a set of connected lanes, encompassing predecessor lanes, successor lanes, and their interconnections.
This unified representation of lanes and lane topology achieves a unified perception scheme, enabling the model to simultaneously detect lane positions and predict lane topological relationships.
As shown in Figure~\ref{fig:compare}, our model captures the connectivity relationships between piecewise lanes by defining connection queries, differing from the traditional reasoning-by-detection paradigm.

Specifically, UniTopo defines two groups of queries for piecewise lanes and connected lanes, uses a shared lane decoder to interact with BEV features, and employs a shared lane head to obtain lane positions and the lane-to-lane topology relationships. 
In addition, we design a Topology-Aware Attention Module~(TAM) to incorporate lane connection information into the features of piecewise lanes.
We use a separate Deformable DETR~\cite{DeformableDETR} to detect traffic elements and adopt an MLP-based topology head to output the topological relationships between lanes and traffic elements.

Our contributions are summarized as threefold:
\begin{itemize}
\item We propose a method for unified modeling of lane and lane topology that concurrently perceives lanes and their topological structures, establishing a new paradigm distinct from the reasoning-by-detection approach.
\item We design a representation that depicts lane-to-lane topology as a set of connected lanes, allowing us to represent lane topology in the same manner as piecewise lanes, thereby achieving unified modeling of lane and lane topology.
\item 
We validate our method on the driving scene reasoning benchmark OpenLane-V2~\cite{OpenLane-V2}, which consists of two subsets, built based on Argoverse2 and nuScenes, respectively.
UniTopo achieves $\text{TOP}_{ll}$ of 30.1\% and 31.8\% on the two subsets, significantly surpassing the existing state-of-the-art method $\text{T}^2\text{SG}$ by 6.0\% and 8.6\%.
\end{itemize}
\section{Related Work}
\subsection{Lane Perception}
Lane perception is a critical component of modern autonomous driving systems.
The objective of 2D lane detection~\cite{linecnn, laneatt, clrnet, CLRNetV2, DHPM, smfrnet, vil, RVLD, STADet, lanetca} is to identify the precise location of lanes in 2D images.
CLRNet~\cite{clrnet, CLRNetV2} introduces a novel network architecture to fully utilize low-level and high-level features.
DHPM~\cite{DHPM} presents a dense hybrid proposal
modulation method to improve the overall quality of
lane proposals.
SMFRNet~\cite{smfrnet} proposes a multi-dimensional feature refinement method for complex scene lane detection based on start point guidance. 
Most existing 2D lane detection methods primarily focus on individual frames and often neglect temporal dynamics across video sequences.
To address this limitation, VIL-100~\cite{vil} constructs a large-scale video instance lane detection dataset.
Following this, STADet~\cite{STADet} presents a streaming video lane detection training framework that explicitly models temporal flow to improve video-based lane perception.
LaneTCA~\cite{lanetca} designs an accumulative attention module and an adjacent attention module to abstract the long-term and short-term temporal context, respectively.
However, these methods are confined to 2D image space and do not incorporate 3D geometric information, which limits their applicability in real-world autonomous driving scenarios. 

Building upon 2D lane detection, 3D lane detection predicts the 3D spatial coordinates of lanes.
Early monocular 3D lane detection methods~\cite{3dlanenet, genlanenet, 3dlanenet+, CLGo} employed Inverse Perspective Mapping~(IPM) to project 2D features into 3D space. However, this approach relies on the assumption of a flat ground, which leads to errors in lane positioning.
Subsequently, to achieve more robust detection results, PersFormer~\cite{PersFormer} uses deformable attention~\cite{DeformableDETR} to construct BEV features and performs lane positioning within the BEV space. Anchor3DLane~\cite{Anchor3DLane, Anchor3DLane++} defines anchors or queries in 3D space and then predicts 3D lanes based on sampled 2D image features.
LATR~\cite{LATR} introduces a lane-aware query generator and dynamic 3D ground positional embedding, directly predicting 3D lanes from image features through cross-attention.

With the advancement of higher-level assisted driving functions, the need for lane perception has progressively transitioned from 3D lane detection to the construction of multi-view online High-Definition~(HD) maps.
Previous studies~\cite{vpn, cvt, gkt} conducted semantic segmentation on BEV features to construct rasterized HD maps. In contrast, recent methods~\cite{HDMapNet, VectorMapNet, MapTR, MapTRv2, MapQR} organize HD maps into vectorized formats that are more suitable for downstream tasks.
HDMapNet~\cite{HDMapNet} vectorizes BEV rasterized HD maps using cluster instance embeddings.
VectorMapNet~\cite{VectorMapNet} employs a two-stage network that first predicts the key points of map elements and then generates ordered point sets through a polyline generator, achieving end-to-end vectorized HD map construction.
MapTR~\cite{MapTR} and MapTRv2~\cite{MapTRv2} model each map element as a set of equivalently arranged points, simplifying the map element prediction task into a parallel regression problem.
MapQR~\cite{MapQR} introduces a scatter-and-gather query design, which scatters instance queries into hierarchical queries through position embedding, thereby reducing computation in the self-attention mechanism.
StreamMapNet~\cite{streammapnet} employs multi-point attention and temporal information to construct large-range and long-sequence local HD maps.

\subsection{Lane Topology Reasoning}
The task of lane topology reasoning aims to predict the connectivity relationships between lanes segments.
As a pioneering work, STSU~\cite{STSU} follows a DETR-like~\cite{DETR} framework, using queries to detect lanes and objects, and then encoding lane queries to perform topology prediction.
CenterLineDet~\cite{CenterLineDet} extends the input modalities to include multi-view images and LiDAR, and employs imitation learning to obtain a centerline graph.
The introduction of the OpenLane-V2~\cite{OpenLane-V2} dataset has heightened attention to lane topology reasoning in driving scenes.
TopoNet~\cite{TopoNet} utilizes GCN~\cite{GCN} to model topological relationships in lane detection. 
TopoMLP~\cite{TopoMLP} proposes a first-detect-then-reason philosophy for better topology prediction.
Topo2D~\cite{Topo2D} enhances 3D lane detection and topology reasoning using 2D lane priors. 
LaneGAP~\cite{LaneGAP} adopts a two-stage strategy by first predicting lane paths and then applying post-processing to derive piecewise lane segments and their topology.
CGNet~\cite{CGNet} tackles lane discontinuities by designing specialized modules, and RoadPainter~\cite{roadpainter} improves centerline accuracy by extracting representative points from each centerline mask. 
TSTGT~\cite{TSTGT} develops a divide-and-conquer topology graph Transformer that separately models lane-to-lane and lane-to-traffic relationships. 
TopoLogic~\cite{TopoLogic} explicitly maps the geometric distances between lanes into topology relationships.
RATopo~\cite{ratopo} proposes a redundancy assignment strategy for lane topology reasoning that enables quantity-rich and geometry-diverse topology supervision.
Another line of work~\cite{SafeDriveRAG, TopoFormer, SGFormer, qi2019attentive} formulates lane topology reasoning as a scene graph construction problem. 
For instance, $\text{T}^2\text{SG}$~\cite{TopoFormer} introduces a Traffic Topology Scene Graph that provides a more accurate and interpretable representation of topological relationships in traffic scenes. 

Compared to these methods, in this paper we propose a method for unified modeling of lane and lane topology. Our method represents the lane-to-lane topology as a set of connected lanes, enabling the direct perception of lane-to-lane topology from image features.
\section{Method}

\begin{figure*}[t]
  \centering
  \includegraphics[width=1.0\linewidth]{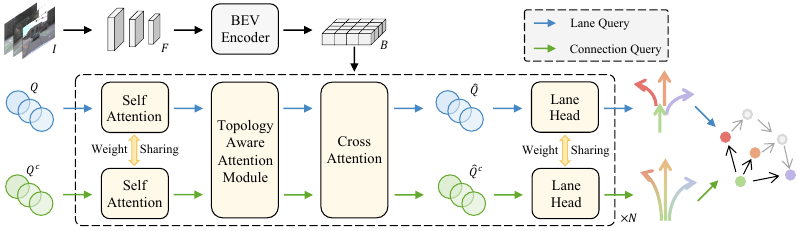}
  \caption{\textbf{The overall architecture of our UniTopo.}
  Within the image feature extractor, the multi-view images are processed through a backbone network (e.g., ResNet-50~\cite{ResNet}) and a neck network (e.g., FPN~\cite{FPN}) to extract features.
  These image features are then transformed into BEV features via a BEV encoder following BEVFormer~\cite{BEVFormer}, where the BEV representation serves as an intermediate feature rather than an input modality. 
  In the lane detector, we define lane queries for detecting piecewise lanes and connection queries for detecting connected lanes.
  We additionally incorporate a Topology-Aware Attention Module~(TAM) to introduce the lane topology information into lane features.
  The topology prediction heads are implemented using MLPs.
  For simplicity, we omit the traffic element detector and the topology prediction head that predicts the topology relationships between lanes and traffic elements.
  }
  \label{fig:fig_3_overall_architecture}
\end{figure*}

\subsection{Problem Definition}
Given multi-view images $\mathbf{I}\in\mathbb{R}^{N_I\times H_I\times W_I\times C}$ captured from $N_I$ cameras, the objective of the topology reasoning task is to simultaneously perform lane centerline detection, traffic element detection, lane-to-lane topology prediction, and lane-to-traffic element topology prediction within end-to-end networks.
Each centerline is represented as a set of uniformly sampled discrete points $\{\mathbf{L}_i\in\mathbb{R}^{N_P\times 3}|i=1,2,...,N_L\}$, where $N_L$ is the number of centerline instances, and $N_P$ is the number of sampling points per centerline.
The detection of 2D traffic elements is performed only on the front view. Each 2D traffic element is represented as a bounding box $\{\mathbf{T}_i\in\mathbb{R}^{4}|i=1,2,...,N_T\}$, where $N_T$ is the number of traffic element instances.
The topology relationships in the scene are represented by two matrices. Specifically, $\mathbf{G}^{ll}\in\mathbb{R}^{N_L\times N_L}$ indicates whether lanes $\mathbf{L}_i$ and $\mathbf{L}_j$ are connected, and $\mathbf{G}^{lt}\in\mathbb{R}^{N_L\times N_T}$ denotes whether lane $\mathbf{L}_i$ and traffic element $\mathbf{T}_j$ are related.
Given two lanes $\mathbf{L}_i\in\mathbb{R}^{N_P\times 3}$ and $\mathbf{L}_j\in\mathbb{R}^{N_P\times 3}$ with a topology relationship, a connected lane $\mathbf{L}^{c}_{i,j}\in\mathbb{R}^{(2N_P-1)\times 3}$ is created by merging them at the junction point, which serves as the terminal point of $\mathbf{L}_i$ and the initial point of $\mathbf{L}_j$.
The merged lane is then resampled to obtain $\mathbf{L}^{c}_{i,j}\in\mathbb{R}^{N_P\times 3}$. 

\subsection{Overview}
Our UniTopo comprises the following components: an image feature extractor, a BEV feature encoder, a lane detector, a traffic element detector, and two topology prediction heads.
For simplicity, in Figure~\ref{fig:fig_3_overall_architecture}, we omit the traffic element detector and the topology prediction head responsible for predicting the topology relationships between lanes and traffic elements.
In the image feature extractor, the multi-view images are processed through a backbone network (e.g., ResNet-50~\cite{ResNet}) and a neck network (e.g., FPN~\cite{FPN}) to extract features $\mathbf{F}\in\mathbb{R}^{N_I\times H_F\times W_F\times C}$.
Subsequently, the BEV features $\mathbf{B}\in\mathbb{R}^{H_B\times W_B\times C}$ are constructed following the approach of BEVFormer~\cite{BEVFormer}.
Both the lane detector and the traffic element detector are designed based on Deformable DETR~\cite{DeformableDETR}.
Additionally, in the lane detector, we incorporate a Topology-Aware Attention Module~(TAM) to embed lane topology information into the lane features.
The topology prediction heads for both lane-to-lane and lane-to-traffic element relationships are implemented using MLPs.
The details will be described in the following sections.

\subsection{Lane Detector}
In the lane detector, we define two groups of queries: lane queries $\{\mathbf{Q}_i\}_{i=1}^{\bar{N}_L}$ for detecting piecewise lanes and connection queries $\{\mathbf{Q}_i^{c}\}_{i=1}^{\bar{N}_L}$ for detecting connected lanes, where $\bar{N}_L$ denotes the maximum number of lanes that can be predicted.
Both lane queries and connection queries are implemented as learnable query embeddings and are randomly initialized.
They are jointly optimized with the rest of the network during end-to-end training, without relying on predefined anchors. 
First, we employ a shared intra-group self-attention module to facilitate feature interaction within each group of queries. Specifically, the updated lane queries are computed as follows:
\begin{gather}
\bar{\mathbf{Q}} = \text{LN}(\text{MultiHeadAttn}(\mathbf{Q}, \mathbf{P}) + \mathbf{Q}),
\end{gather}
where $\mathbf{P}$ is a learnable positional embedding used in Deformable DETR~\cite{DeformableDETR}.
Similarly, we obtain the updated connection queries $\bar{\mathbf{Q}}^{c}$.

Since we model the lane-lane topology as connected lanes, the connection queries encode the topological relationships of the scene.
Therefore, we design a Topology-Aware Attention Module~(TAM) to integrate this lane connection information into the lane features:
\begin{gather}
\tilde{\mathbf{Q}} = \text{TAM}(\bar{\mathbf{Q}}, \bar{\mathbf{Q}}^{c}; \{\bar{\mathbf{L}}_i\}_{i=1}^{\bar{N}_L}, \{\bar{\mathbf{L}}^{c}_i\}_{i=1}^{\bar{N}_L}), \\
\tilde{\mathbf{Q}}^c = \bar{\mathbf{Q}}^c,
\end{gather}
where $\{\bar{\mathbf{L}}_i\}_{i=1}^{\bar{N}_L}$ and $\{\bar{\mathbf{L}}^{c}_i\}_{i=1}^{\bar{N}_L}$ denote the lane predictions from the previous decoder layer.

Subsequently, $\tilde{\mathbf{Q}}$ aggregates the BEV features using a deformable attention module and then interacts with the associated traffic element features via a GCN:
\begin{gather}
\mathbf{V} = \text{LN}(\text{DeformAttn}(\tilde{\mathbf{Q}}, \mathbf{B}) + \tilde{\mathbf{Q}}), \\
\hat{\mathbf{Q}} = \text{LN}(\text{GCN}(\text{FFN}(\mathbf{V}), \mathbf{Q}^t; \bar{\mathbf{G}}^{ll}, \bar{\mathbf{G}}^{lt}) + \mathbf{V}),
\end{gather}
where $\bar{\mathbf{G}}^{ll}$ and $\bar{\mathbf{G}}^{lt}$ denote the topology predictions from the previous decoder layer, and $\mathbf{Q}^t$ represents the traffic element features from the traffic element detector.
We obtain $\hat{\mathbf{Q}}^c$ in the same way.

Finally, we obtain the lane prediction results using a lane head that consists of a classification branch and a regression branch.
The classification branch estimates the existence probability of each lane instance, while the regression branch predicts the 3D coordinates of the lane points.
To supervise the predicted lanes $\bar{\mathbf{L}}$, we utilize the vanilla ground truths $\{\mathbf{L}_i\in\mathbb{R}^{N_P\times 3}|i=1,2,...,N_L\}$.
For the connected lanes $\bar{\mathbf{L}}^c$, we employ the constructed ground truths for supervision, defined as $\{\mathbf{L}^c_{i,j}\in\mathbb{R}^{N_P\times 3}|i=1,2,...,N_L;j=1,2,...,N_L;\mathbf{G}^{ll}_{i,j}=1\}$.

\subsection{Topology-Aware Attention Module}

\begin{figure}[t]
  \centering
  \includegraphics[width=1.0\linewidth]{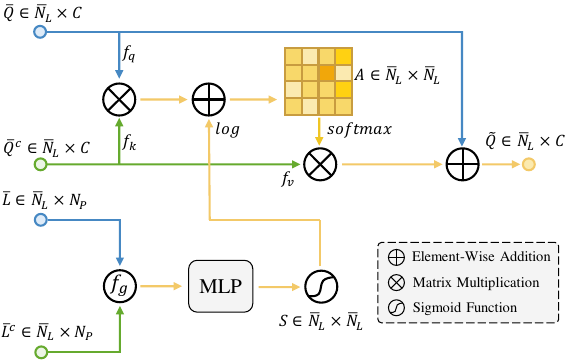}
  \caption{\textbf{Illustration of the Topology-Aware Attention Module.} First, the correlation between piecewise lanes and connected lanes is measured based on geometric distance. Then, utilizing a cross-attention mechanism, the lane topology information contained in the connection queries is transferred to the lane features.
  }
  \label{fig:fig_4_TAM}
\end{figure}

As illustrated in Figure~\ref{fig:fig_4_TAM}, the Topology-Aware Attention Module~(TAM) operates in two steps. First, we measure the correlation between piecewise lanes and connected lanes based on geometric distance. Then, using a cross-attention mechanism, we transfer the lane topology information contained in the connection queries to the piecewise lane features.

Specifically, to compute the correlation between each predicted piecewise lane and each connected lane, we calculate the average L1 distances between $\bar{\mathbf{L}}$ and both the first and the second half of $\bar{\mathbf{L}}^{c}$. 
We select the smaller of the two distances as the correlation measure:
\begin{gather}
\mathbf{D} = \text{Min}(f_g(\bar{\mathbf{L}}, \bar{\mathbf{L}}^{c}_1),f_g(\bar{\mathbf{L}}, \bar{\mathbf{L}}^{c}_2)),
\end{gather}
where $f_g(\cdot)$ denotes the average L1 distance, $\bar{\mathbf{L}}^{c}_1$ and $\bar{\mathbf{L}}^{c}_2$ represent the first and second halves of $\bar{\mathbf{L}}^{c}$. The correlation score $\mathbf{S}$ is then computed as:
\begin{gather}
\mathbf{S} = \text{Sigmoid}(\text{MLP}(\mathbf{D})).
\end{gather}
Next, we employ the lane features as queries and the connection features as keys and values, using $\mathbf{S}$ as the attention mask. 
To incorporate the topology information into the lane features, we perform cross-attention defined as:
\begin{gather}
\mathbf{A} = \text{Softmax}(\frac{f_q(\bar{\mathbf{Q}})\otimes f_k(\bar{\mathbf{Q}}^{c})^T}{\sqrt{C}} + \log(\mathbf{S})), \\
\tilde{\mathbf{Q}} = \text{LN}(\mathbf{A}\otimes f_v(\bar{\mathbf{Q}}^{c}) + \bar{\mathbf{Q}}),
\end{gather}
where $f_q(\cdot)$, $f_k(\cdot)$, and $f_v(\cdot)$ are linear transformations applied to the input features, and $\otimes$ denotes matrix multiplication.

\subsection{Topology Prediction Head}
By modeling the lane-to-lane topology as connected lanes and aggregating their features from images, connection queries can be directly utilized to infer the topological relationships between lanes.
Since the prediction of topological relationship needs to correspond with the perception results, for each connected lane, we identify two piecewise lanes that have the smallest average L1 distances to its first and second halves, respectively.
Formally, the indices of the piecewise lanes are defined as:
\begin{gather}
i^\ast = \arg\min_{i} f_g(\bar{\mathbf{L}}_i, \bar{\mathbf{L}}^{c}_1), \\
j^\ast = \arg\min_{i} f_g(\bar{\mathbf{L}}_j, \bar{\mathbf{L}}^{c}_2).
\end{gather}
where $f_g(\cdot)$ denotes the average L1 distance.
The features of the two selected piecewise lanes are represented as $\hat{\mathbf{Q}}_{i^\ast}$ and $\hat{\mathbf{Q}}_{j^\ast}$, while the feature of the corresponding connected lane is represented as $\hat{\mathbf{Q}}_{i^\ast,j^\ast}^c$.
For simplicity, we denote $i^\ast$ and $j^\ast$ as $i$ and $j$ in the following paragraphs. 

Based on these features, the connection relationship between these two piecewise lanes is predicted as follows:
\begin{gather}
\mathbf{Q}^{ll}_{i,j} = \text{Concat}(\text{MLP}(\hat{\mathbf{Q}}^{c}_{i,j}+\hat{\mathbf{Q}}_i), \text{MLP}(\hat{\mathbf{Q}}^{c}_{i,j}+\hat{\mathbf{Q}}_j)), \\
\bar{\mathbf{G}}^{ll}_{i,j} = \text{Sigmoid}(\text{MLP}(\mathbf{Q}^{ll}_{i,j})).
\end{gather}

For piecewise lane pairs that are not matched to any connected lanes, $\mathbf{Q}^{ll}_{i,j}$ is defined as:
\begin{gather}
\mathbf{Q}^{ll}_{i,j} = \text{Concat}(\text{MLP}(\hat{\mathbf{Q}}_i), \text{MLP}(\hat{\mathbf{Q}}_j)).
\end{gather}

The prediction of topology relationships between lanes and traffic elements is based solely on feature similarity:
\begin{gather}
\mathbf{Q}^{lt}_{i,j} = \text{Concat}(\text{MLP}(\hat{\mathbf{Q}}_i), \text{MLP}(\mathbf{Q}^t_j)), \\
\bar{\mathbf{G}}^{lt}_{i,j} = \text{Sigmoid}(\text{MLP}(\mathbf{Q}^{lt}_{i,j})),
\end{gather}
where $\mathbf{Q}^t_j$ denotes the features of the $j^{th}$ traffic element instance.

\begin{table*}[t]
  \caption{\textbf{Comparison results of lane centerline perception and reasoning on OpenLane-V2~\cite{OpenLane-V2} \textit{subset\_A}}.
  $*$: The results are re-implemented by us based on the open-source code of TopoMLP~\cite{TopoMLP}.
  $\dagger$: The results for TSTGT$\dagger$~\cite{TSTGT} are cited from their paper, which uses outdated metrics due to the unavailability of open-source code.
  $\ddagger$: UniTopo is built upon the well-recognized baseline TopoNet~\cite{TopoNet}, while UniTopo$\ddagger$ is implemented with TopoLogic~\cite{TopoLogic} as the baseline.
  }
  \label{tab:sota_subseta}
  \centering
  \renewcommand\arraystretch{1.2}
  \begin{tabular}{c|c|c|c|ccccc}
    \thickhline
    Method & Venues & Backbone & Epoch & $\text{TOP}_{ll}\uparrow$ & $\text{TOP}_{lt}\uparrow$ & $\text{DET}_l\uparrow$ & $\text{DET}_t\uparrow$ & $\text{OLS}\uparrow$ \\
    \hline \hline
    STSU~\cite{STSU} & ICCV 2021 & ResNet-50 & 24e & 2.9 & 19.8 & 12.7 & 43.0 & 29.3 \\
    VectorMapNet~\cite{VectorMapNet} & ICML 2023 & ResNet-50 & 24e & 2.7 & 9.2 & 11.1 & 41.7 & 24.9 \\
    MapTR~\cite{MapTR} & ICLR 2023 & ResNet-50 & 24e & 5.9 & 15.1 & 17.7 & 43.5 & 31.0 \\
    TopoNet~\cite{TopoNet} & Arxiv 2023 & ResNet-50 & 24e & 10.9 & 23.8 & 28.6 & 48.6 & 39.8 \\
    TopoMLP~\cite{TopoMLP} & ICLR 2024 & ResNet-50 & 24e & 21.7 & 26.9 & 28.5 & 49.5 & 44.1 \\
    Topo2D~\cite{Topo2D} & Arxiv 2024 & ResNet-50 & 24e & 22.3 & 26.2 & 29.1 & 50.6 & 44.5 \\
    RoadPainter~\cite{roadpainter} & ECCV 2024 & ResNet-50 & 24e & 22.8 & 27.2 & 30.7 & 47.7 & 44.6 \\
    TSTGT$\dagger$~\cite{TSTGT} & ACM MM 2024 & ResNet-50 & 24e & 12.1 & 23.5 & 29.0 & 50.5 & 40.7 \\
    TopoLogic~\cite{TopoLogic} & NeurIPS 2024 & ResNet-50 & 24e & 23.9 & 25.4 & 29.9 & 47.2 & 44.1 \\
    $\text{T}^2\text{SG}$~\cite{TopoFormer} & CVPR 2025 & ResNet-50 & 24e & 24.1 & 29.5 & \textbf{34.7} & 48.2 & 46.3 \\
    \cellcolor{title_gray}UniTopo (Ours) & \cellcolor{title_gray}- & \cellcolor{title_gray}ResNet-50 & \cellcolor{title_gray}24e & \cellcolor{title_gray}29.6 & \cellcolor{title_gray}31.2 & \cellcolor{title_gray}29.8 & \cellcolor{title_gray}49.3 & \cellcolor{title_gray}47.3 \\
    \cellcolor{title_gray}UniTopo$\ddagger$ (Ours) & \cellcolor{title_gray}- & \cellcolor{title_gray}ResNet-50 & \cellcolor{title_gray}24e & \cellcolor{title_gray}\textbf{30.1} & \cellcolor{title_gray}\textbf{31.4} & \cellcolor{title_gray}31.4 & \cellcolor{title_gray}\textbf{52.4} & \cellcolor{title_gray}\textbf{48.7} \\
    \cline{1-9}
    TopoMLP$*$~\cite{TopoMLP} & ICLR 2024 & Swin-B & 24e & 24.7 & 31.5 & 31.1 & \textbf{55.3} & 48.1 \\
    TopoMLP$*$~\cite{TopoMLP} & ICLR 2024 & Swin-B & 48e & 28.7 & 33.5 & 32.9 & 54.4 & 49.7 \\
    \cellcolor{title_gray}UniTopo$\ddagger$ (Ours) & \cellcolor{title_gray}- & \cellcolor{title_gray}Swin-B & \cellcolor{title_gray}24e & \cellcolor{title_gray}31.5 & \cellcolor{title_gray}33.8 & \cellcolor{title_gray}31.6 & \cellcolor{title_gray}52.4 & \cellcolor{title_gray}49.5 \\
    \cellcolor{title_gray}UniTopo$\ddagger$ (Ours) & \cellcolor{title_gray}- & \cellcolor{title_gray}Swin-B & \cellcolor{title_gray}48e & \cellcolor{title_gray}\textbf{35.1} & \cellcolor{title_gray}\textbf{35.8} & \cellcolor{title_gray}\textbf{34.2} & \cellcolor{title_gray}50.8 & \cellcolor{title_gray}\textbf{51.0} \\
    \hline
  \end{tabular}
\end{table*}

\subsection{Loss Function}
We define the overall loss function of our UniTopo as follows: 
\begin{gather}
\mathcal{L} = \lambda^{l}\mathcal{L}^{l} + \lambda^{t}\mathcal{L}^{t} + \lambda^{ll}\mathcal{L}^{ll} + \lambda^{lt}\mathcal{L}^{lt},
\end{gather}
where $\mathcal{L}^{l}$, $\mathcal{L}^{t}$, $\mathcal{L}^{ll}$, and $\mathcal{L}^{lt}$ denote the losses associated with lane detection, traffic element detection, lane-to-lane topology prediction, and lane-to-traffic element topology prediction, respectively.
The lane detection loss $\mathcal{L}^{l}$ consists of two components, expressed as follows:
\begin{gather}
\mathcal{L}^{l} = \lambda^{l}_{cls}\mathcal{L}_{cls} + \lambda^{l}_{reg}\mathcal{L}_{reg},
\end{gather}
where $\mathcal{L}_{cls}$ represents the focal loss~\cite{focal}, and $\mathcal{L}_{reg}$ denotes the L1 loss. 
In addition to the classification loss and regression loss, the traffic element detection loss $\mathcal{L}^{t}$ also includes the GIoU loss~\cite{giou}:
\begin{gather}
\mathcal{L}^{t} = \lambda^{t}_{cls}\mathcal{L}_{cls} + \lambda^{t}_{reg}\mathcal{L}_{reg} + \lambda^{t}_{iou}\mathcal{L}_{iou},
\end{gather}
For topology reasoning, we adopt the focal loss~\cite{focal} for both $\mathcal{L}^{ll}$ and $\mathcal{L}^{lt}$.

\subsection{Group-based Topology Training Strategy}
Inspired by the one-to-many label assignment commonly used in current object detection methods~\cite{FasterRCNN,GroupDETR,HDETR}, we propose a group-based topology training strategy tailored for topology reasoning tasks to accelerate the convergence of the topology head.
Specifically, we define two architectural query types: a lane query group $\{\mathbf{Q}_i\}_{i=1}^{\bar{N}_L}$ and a connection query group $\{\mathbf{Q}_i^{c}\}_{i=1}^{\bar{N}_L}$, denoted as $\mathbf{G}$ and $\mathbf{G}^c$, respectively. These query types are part of the model architecture and are shared across training and inference.
During training, we replicate each of the two query groups $K$ times, resulting in $K$ lane query groups $\{\mathbf{G}_1, \mathbf{G}_2, \ldots, \mathbf{G}_K\}$ and $K$ connection query groups $\{\mathbf{G}^c_1, \mathbf{G}^c_2, \ldots, \mathbf{G}^c_K\}$.
For each query group, a one-to-one assignment between predictions and ground-truth lanes is performed using Hungarian matching, with the matching cost defined by the same classification and regression loss terms used for training.
Matched query-ground-truth pairs are supervised accordingly, while unmatched queries are treated as negatives.
This matching and supervision process is applied independently to each query group. 

Unlike prior detection-oriented one-to-many methods, our strategy not only allows each lane query to encode detection-related attributes (e.g., position, category) but also enables each connection query to capture topology-specific cues (e.g., connectivity), thereby enhancing the supervision for predicting topological relationships.
\section{Experiments}

\begin{table*}[t]
  \caption{\textbf{Comparison results of lane centerline perception and reasoning on OpenLane-V2~\cite{OpenLane-V2} \textit{subset\_B}}.
  $*$: The results are re-implemented by us based on the open-source code of TopoMLP~\cite{TopoMLP}.
  $\dagger$: The results for TSTGT$\dagger$~\cite{TSTGT} are cited from their paper, which uses outdated metrics due to the unavailability of open-source code.
  $\ddagger$: UniTopo is built upon the well-recognized baseline TopoNet~\cite{TopoNet}, while UniTopo$\ddagger$ is implemented with TopoLogic~\cite{TopoLogic} as the baseline.
  }
  \label{tab:sota_subsetb}
  \centering
  \renewcommand\arraystretch{1.2}
  \begin{tabular}{c|c|c|c|ccccc}
    \thickhline
    Method & Venues & Backbone & Epoch & $\text{TOP}_{ll}\uparrow$ & $\text{TOP}_{lt}\uparrow$ & $\text{DET}_l\uparrow$ & $\text{DET}_t\uparrow$ & $\text{OLS}\uparrow$ \\
    \hline \hline
    TopoNet~\cite{TopoNet} & Arxiv 2023 & ResNet-50 & 24e & 6.7 & 16.7 & 24.4 & 52.6 & 36.0 \\
    TopoMLP$*$~\cite{TopoMLP} & ICLR 2024 & ResNet-50 & 24e & 21.6 & 20.0 & 25.8 & 60.2 & 44.3 \\
    TSTGT$\dagger$~\cite{TSTGT} & ACM MM 2024 & ResNet-50 & 24e & 13.7 & 18.9 & 27.5 & \textbf{60.5} & 42.1 \\
    TopoLogic~\cite{TopoLogic} & NeurIPS 2024 & ResNet-50 & 24e & 21.6 & 17.9 & 25.9 & 54.7 & 42.3 \\
    $\text{T}^2\text{SG}$~\cite{TopoFormer} & CVPR 2025 & ResNet-50 & 24e & 23.2 & 23.3 & \textbf{34.8} & 58.9 & 47.5 \\
    \cellcolor{title_gray}UniTopo (Ours) & \cellcolor{title_gray}- & \cellcolor{title_gray}ResNet-50 & \cellcolor{title_gray}24e & \cellcolor{title_gray}28.3 & \cellcolor{title_gray}23.4 & \cellcolor{title_gray}29.4 & \cellcolor{title_gray}58.6 & \cellcolor{title_gray}47.4 \\
    \cellcolor{title_gray}UniTopo$\ddagger$ (Ours) & \cellcolor{title_gray}- & \cellcolor{title_gray}ResNet-50 & \cellcolor{title_gray}24e & \cellcolor{title_gray}\textbf{31.8} & \cellcolor{title_gray}\textbf{24.3} & \cellcolor{title_gray}31.1 & \cellcolor{title_gray}53.6 & \cellcolor{title_gray}\textbf{47.6} \\
    \cline{1-9}
    TopoMLP${*}$~\cite{TopoMLP} & ICLR 2024 & Swin-B & 24e & 27.4 & 26.1 & 30.5 & 66.8 & 50.2 \\
    TopoMLP${*}$~\cite{TopoMLP} & ICLR 2024 & Swin-B & 48e & 33.1 & 27.4 & 37.2 & \textbf{67.1} & 53.6 \\
    \cellcolor{title_gray}UniTopo$\ddagger$ (Ours) & \cellcolor{title_gray}- & \cellcolor{title_gray}Swin-B & \cellcolor{title_gray}24e & \cellcolor{title_gray}34.4 & \cellcolor{title_gray}27.7 & \cellcolor{title_gray}33.1 & \cellcolor{title_gray}61.0 & \cellcolor{title_gray}51.4 \\
    \cellcolor{title_gray}UniTopo$\ddagger$ (Ours) & \cellcolor{title_gray}- & \cellcolor{title_gray}Swin-B & \cellcolor{title_gray}48e & \cellcolor{title_gray}\textbf{40.7} & \cellcolor{title_gray}\textbf{30.7} & \cellcolor{title_gray}\textbf{37.4} & \cellcolor{title_gray}62.0 & \cellcolor{title_gray}\textbf{54.6} \\
    \hline
  \end{tabular}
\end{table*}

\begin{table}[t]
  \caption{
  \textbf{Comparison results of lane segment perception and reasoning on the OpenLane-V2~\cite{OpenLane-V2} \textit{subset\_A}}.
  \textit{Subset\_B} is not evaluated because it does not provide lane-segment annotations.
  All experiments are conducted using ResNet-50~\cite{ResNet} as the backbone, and are also trained for 24 epochs. 
  }
  \label{tab:laneseg}
  \tabcolsep=0.12cm
  \renewcommand\arraystretch{1.2}
  \centering
  \begin{tabular}{c|c|cccc}
    \thickhline
    Method & Venues & $\text{TOP}_{lsls}\uparrow$ & $\text{mAP}\uparrow$ & $\text{AP}_{ls}\uparrow$ & $\text{AP}_{ped}\uparrow$ \\
    \hline \hline
    TopoNet~\cite{TopoNet} & Arxiv 2023 & - & 23.0 & 23.9 & 22.0 \\
    MapTR~\cite{MapTR} & ICLR 2023 & - & 27.0 & 25.9 & 28.1 \\
    MapTRv2~\cite{MapTRv2} & IJCV 2024 & - & 28.5 & 26.6 & 30.4 \\
    LaneSegNet~\cite{LaneSegNet} & ICLR 2024 & 25.4 & 32.6 & 32.3 & 32.9 \\
    \rowcolor{title_gray}
    UniTopo~(Ours) & \cellcolor{title_gray}- & \textbf{31.4} & \textbf{34.4} & \textbf{32.7} & \textbf{36.1} \\
    \hline
  \end{tabular}
\end{table}

\begin{table}[t]
  \caption{\textbf{Effectiveness of different modules.} 
  \textit{UniQuery} denotes unified piecewise lane queries and connection queries.
  \textit{Groups} denotes the group-based topology training strategy.
  \textit{GeoDist} denotes the method proposed by TopoLogic~\cite{TopoLogic}.
  }
  \label{tab:module}
  \renewcommand\arraystretch{1.2}
  \centering
  \begin{tabular}{c|ccccc}
    \thickhline
    Method & $\text{TOP}_{ll}\uparrow$ & $\text{TOP}_{lt}\uparrow$ & $\text{DET}_l\uparrow$ & $\text{DET}_t\uparrow$ & $\text{OLS}\uparrow$ \\
    \hline \hline
    TopoNet & 10.9 & 23.8 & 28.6 & 48.6 & 39.8 \\
    Baseline & 12.4 & 26.0 & 30.0 & 50.4 & 41.6 \\
    + UniQuery & 23.3 & 27.9 & 29.4 & 49.4 & 45.0 \\
    + TAM & 25.0 & 28.8 & 30.2 & 51.1 & 46.2 \\
    + Groups & 29.6 & 31.2 & 29.8 & 49.3 & 47.3 \\
    + GeoDist & \textbf{30.1} & \textbf{31.4} & \textbf{31.4} & \textbf{52.4} & \textbf{48.7} \\
    \hline
  \end{tabular}
\end{table}

\subsection{Datasets and Metrics}
To evaluate our proposed method, we conduct experiments on the topology reasoning benchmark OpenLane-V2~\cite{OpenLane-V2}. OpenLane-V2~\cite{OpenLane-V2} comprises two subsets, \textit{subset\_A} and \textit{subset\_B}, which are annotated based on the Argoverse2~\cite{Argoverse2} and nuScenes~\cite{nuScenes} datasets, respectively. Each subset contains 1000 scenes with annotations for lane centerlines, traffic elements, lane-to-lane topology, and lane-to-traffic element topology. \textit{Subset\_A} provides seven camera views per frame, whereas \textit{subset\_B} offers six camera views per frame.

The evaluation metrics consist of four components, all based on mean Average Precision (mAP). The metrics $\text{DET}_l$ and $\text{DET}_t$ assess the detection performance of lane centerlines and traffic elements, respectively. Meanwhile, $\text{TOP}_{ll}$ and $\text{TOP}_{lt}$ evaluate the topology reasoning performance for lane-to-lane and lane-to-traffic element relationships. Specifically, $\text{DET}_l$ utilizes the Frech{\'e}t distance to measure the similarity between lanes, averaging the results at thresholds of $1.0$m, $2.0$m, and $3.0$m. In contrast, $\text{DET}_t$ employs the Intersection over Union~(IoU) metric and computes the average results across different categories of traffic elements. The metrics $\text{TOP}_{ll}$ and $\text{TOP}_{lt}$ are specifically designed for graph data and are measured based on the matching results between predicted lanes, traffic elements, and the ground truths. The overall OpenLane-V2 Score (OLS) is calculated as follows:
\begin{equation}
  \text{OLS} = \frac{1}{4}[f(\text{TOP}_{ll})+f(\text{TOP}_{lt})+\text{DET}_l+\text{DET}_t],
  \label{eq:OLS}
\end{equation}
where $f$ is a scale function to emphasize the task of topology reasoning.

Building upon centerline perception, Li et al.~\cite{LaneSegNet} propose a novel mapping formulation known as lane segment perception. They also re-annotated the \textit{subset\_A} of OpenLane-V2~\cite{OpenLane-V2} in a lane segment manner. Lane segments define the geometric boundaries and preserve the directed topological connections necessary for constructing a lane graph. Lane segment perception utilizes specially designed metrics, including average precision metrics $\text{AP}_{ls}$ and $\text{AP}_{ped}$, along with the mean average precision~(mAP) computed as the average of $\text{AP}_{ls}$ and $\text{AP}_{ped}$. Additionally, it incorporates the topological metric $\text{TOP}_{lsls}$. 
The formal definition of the lane segment distance between the prediction $\mathcal{V}$ and the ground truth $\tilde{\mathcal{V}}$, denoted as $\mathcal{D}_{ls}(\mathcal{V}, \tilde{\mathcal{V}})$, is provided as follows:
\begin{gather}
\mathcal{D}_{lr}(\mathcal{V}, \tilde{\mathcal{V}})=\text{Chamfer}([\mathcal{V}_{l}, \mathcal{V}_{r}], [\tilde{\mathcal{V}}_{l}, \tilde{\mathcal{V}}_{r}]), \\
\mathcal{D}_{c}(\mathcal{V}, \tilde{\mathcal{V}})=\text{Frech{\'e}t}(\mathcal{V}_{c}, \tilde{\mathcal{V}}_{c}), \\
\mathcal{D}_{ls}(\mathcal{V}, \tilde{\mathcal{V}})=\frac{1}{2}\left[\mathcal{D}_{lr}(\mathcal{V}, \tilde{\mathcal{V}}) + \mathcal{D}_{c}(\mathcal{V}, \tilde{\mathcal{V}})\right].
\end{gather}

\subsection{Implementation Details}
For \textit{subset\_A}, all views except the front view have a resolution of $2048 \times 1550$ pixels, while the front view is cropped and padded to match this size. For \textit{subset\_B}, all views are $1600 \times 900$ pixels.
Subsequently, all images are resized by a factor of 0.5.
ResNet-50~\cite{ResNet} is employed as the backbone, and FPN~\cite{FPN} is used as the neck to extract image features at downsampling scales of $4\times$, $8\times$, $16\times$, and $32\times$.
We utilize a 3-layer BEVFormer~\cite{BEVFormer} encoder for view transformation. The BEV features have dimensions of $200\times 100$, corresponding to a perception range from -51.2m to 51.2m along the X-axis, and from -25.6m to 25.6m along the Y-axis.
The parameters $\bar{N}_L$ and $\bar{N}_T$ are set to 300 and 100.
During training, we employ a group-based topology training strategy to accelerate the convergence of the topology head. The number of groups is set to 6.
The loss weights $\lambda^{l}$, $\lambda^{t}$, $\lambda^{ll}$, and $\lambda^{lt}$ are assigned values of 1.0, 1.0, 5.0, and 5.0, respectively.
For lane detection loss, the classification and regression weights are $\lambda^{l}_{cls}=1.5$ and $\lambda^{l}_{reg}=0.025$.
For traffic element detection loss, the classification, regression, and GIoU weights are $\lambda^{t}_{cls}=1.2$, $\lambda^{t}_{reg}=3.0$, and $\lambda^{t}_{iou}=1.2$.
The training is performed using the AdamW~\cite{adamw} optimizer with an initial learning rate of $2\times 10^{-4}$. The batch size is set to 8. All experiments are conducted on 8 Tesla A100 GPUs.

\subsection{Comparison on OpenLane-V2 Dataset}

\begin{table}[t]
  \caption{\textbf{Incorporation of connection queries.}
  \textit{SA} denotes standard self-attention mechanism.
  \textit{GCN} denotes Graph Convolutional Network.
  }
  \label{tab:TAM}
  \renewcommand\arraystretch{1.2}
  \centering
  \begin{tabular}{c|ccccc}
    \thickhline
    Method & $\text{TOP}_{ll}\uparrow$ & $\text{TOP}_{lt}\uparrow$ & $\text{DET}_l\uparrow$ & $\text{DET}_t\uparrow$ & $\text{OLS}\uparrow$ \\
    \hline \hline
    SA & 29.8 & 31.0 & 30.0 & 47.7 & 47.0 \\
    GCN & 28.8 & 30.4 & 30.5 & 48.9 & 47.1 \\
    TAM & \textbf{30.1} & \textbf{31.4} & \textbf{31.4} & \textbf{52.4} & \textbf{48.7} \\
    \hline
  \end{tabular}
\end{table}

\begin{table}[t]
  \caption{
  \textbf{Number of queries.}
  All models are evaluated on OpenLane-V2~\cite{OpenLane-V2} \textit{subset\_A} under the same training and evaluation settings, using the official metrics.
  The second row corresponds to the default configuration used in all other experiments. 
  }
  \label{tab:lane}
  \renewcommand\arraystretch{1.2}
  \centering
  \begin{tabular}{c|ccccc}
    \thickhline
    Number & $\text{TOP}_{ll}\uparrow$ & $\text{TOP}_{lt}\uparrow$ & $\text{DET}_l\uparrow$ & $\text{DET}_t\uparrow$ & $\text{OLS}\uparrow$ \\
    \hline \hline
    200 & 27.0 & 28.9 & 30.4 & 48.5 & 46.2 \\
    300 & 30.1 & 31.4 & \textbf{31.4} & \textbf{52.4} & \textbf{48.7} \\
    400 & \textbf{31.6} & \textbf{32.5} & 30.7 & 48.2 & 48.0 \\
    \hline
  \end{tabular}
\end{table}

\begin{table}[t]
  \caption{
  \textbf{Number of query groups.}
  All models are evaluated on OpenLane-V2~\cite{OpenLane-V2} \textit{subset\_A} under the same training and evaluation settings, using the official metrics.
  The second row corresponds to the default configuration used in all other experiments. 
  }
  \label{tab:group}
  \renewcommand\arraystretch{1.2}
  \centering
  \begin{tabular}{c|ccccc}
    \thickhline
    Number & $\text{TOP}_{ll}\uparrow$ & $\text{TOP}_{lt}\uparrow$ & $\text{DET}_l\uparrow$ & $\text{DET}_t\uparrow$ & $\text{OLS}\uparrow$ \\
    \hline \hline
    4 & 28.4 & 30.1 & 30.3 & 49.0 & 46.9 \\
    6 & \textbf{30.1} & 31.4 & \textbf{31.4} & \textbf{52.4} & \textbf{48.7} \\
    8 & 29.8 & 30.9 & 30.9 & 46.0 & 46.8 \\
    10 & 30.0 & \textbf{31.8} & 30.2 & 44.7 & 46.5 \\
    \hline
  \end{tabular}
\end{table}

\begin{table}[t]
  \caption{\textbf{Loss weight settings.} \textit{Weights} refers to the modified loss weights based on the default settings in TopoNet~\cite{TopoNet}.
  }
  \label{tab:loss}
  \renewcommand\arraystretch{1.2}
  \centering
  \begin{tabular}{c|ccccc}
    \thickhline
    Weights & $\text{TOP}_{ll}\uparrow$ & $\text{TOP}_{lt}\uparrow$ & $\text{DET}_l\uparrow$ & $\text{DET}_t\uparrow$ & $\text{OLS}\uparrow$ \\
    \hline \hline
    - & \textbf{30.1} & \textbf{31.4} & \textbf{31.4} & \textbf{52.4} & \textbf{48.7} \\
    $\lambda^{l}$, $\lambda^{t}$ & 28.2 & 30.1 & 30.2 & 48.3 & 46.6 \\
    $\lambda^{ll}$, $\lambda^{lt}$ & 28.3 & 30.1 & 28.9 & 47.7 & 46.2 \\
    $\lambda^{l}$, $\lambda^{ll}$ & 28.7 & 30.8 & 30.3 & 45.1 & 46.1 \\
    \hline
  \end{tabular}
\end{table}

\begin{table}[t]
  \caption{
  \textbf{BEV dependency analysis.} \textit{Depth} denotes the number of layers in the BEV encoder, and \textit{Resolution} denotes the spatial resolution of the BEV feature map. The last row corresponds to the default configuration used in all other experiments. 
  }
  \label{tab:bev}
  \tabcolsep=0.12cm
  \renewcommand\arraystretch{1.2}
  \centering
  \begin{tabular}{cc|ccccc}
    \thickhline
    Depth & Resolution & $\text{TOP}_{ll}\uparrow$ & $\text{TOP}_{lt}\uparrow$ & $\text{DET}_l\uparrow$ & $\text{DET}_t\uparrow$ & $\text{OLS}\uparrow$ \\
    \hline \hline
    1 & $200\times 100$ & 26.2 & 28.6 & 28.5 & 48.1 & 45.3 \\
    3 & $100\times 50$ & 29.2 & 31.3 & 30.9 & 49.2 & 47.5 \\
    3 & $200\times 100$ & \textbf{30.1} & \textbf{31.4} & \textbf{31.4} & \textbf{52.4} & \textbf{48.7} \\
    \hline
  \end{tabular}
\end{table}

\begin{table}[t]
  \caption{
  \textbf{Design of attention mask.}
  \textit{Min} refers to selecting the minimum of the two distances between the piecewise lane and the first and second halves of the connected lane to compute the attention mask, while \textit{Sum} refers to using the sum of the two distances.
  \textit{Share} indicates whether the distance-mapping MLP is parameter shared across decoder layers. 
  }
  \label{tab:attn}
  \tabcolsep=0.12cm
  \renewcommand\arraystretch{1.2}
  \centering
  \begin{tabular}{cc|ccccc}
    \thickhline
    Distance & Share & $\text{TOP}_{ll}\uparrow$ & $\text{TOP}_{lt}\uparrow$ & $\text{DET}_l\uparrow$ & $\text{DET}_t\uparrow$ & $\text{OLS}\uparrow$ \\
    \hline \hline
    Sum & \checkmark & 24.9 & \textbf{28.9} & 29.8 & 49.2 & 45.7 \\
    Min &  & 24.8 & 28.7 & \textbf{30.3} & 49.5 & 45.8 \\
    Min & \checkmark & \textbf{25.0} & 28.8 & 30.2 & \textbf{51.1} & \textbf{46.2} \\
    \hline
  \end{tabular}
\end{table}

\begin{table}[t]
    \caption{\textbf{Comparison results of OLS and computational cost} on OpenLane-V2 \textit{subset\_A}.
    \textit{Param.} denotes parameters.
    }
    \label{tab:flops}
    \renewcommand\arraystretch{1.2}
    \centering
    \begin{tabular}{c|c|ccc}
    \thickhline
    Method & $\text{TOP}_{ll}\uparrow$ & $\text{FLOPs}\downarrow$ & $\text{Param.}\downarrow$ & $\text{FPS}\uparrow$ \\
    \hline \hline
    TopoNet & 10.9 & 63.07G & 65.96M & 4.57 \\
    UniTopo (Ours) & 29.6 & 77.05G & 67.70M & 4.21 \\
    \hline
    \end{tabular}
\end{table}

Table~\ref{tab:sota_subseta} presents a quantitative comparison on OpenLane-V2 \textit{subset\_A}.
UniTopo is built upon the well-recognized baseline TopoNet. Using ResNet-50 as the backbone and training for 24 epochs, UniTopo achieves a $\text{TOP}_{ll}$ of 29.6\% on \textit{subset\_A}, representing an improvement of 18.7\% over TopoNet.
When implemented with TopoLogic as the baseline, UniTopo$\ddagger$ attains a higher $\text{TOP}_{ll}$ of 30.1\%, significantly outperforming all existing methods, including the state-of-the-art $\text{T}^2\text{SG}$.
When utilizing a larger backbone network (i.e., Swin-B) and extending the training time to 48 epochs, UniTopo$\ddagger$ achieves an improvement of 6.4\% on the $\text{TOP}_{ll}$ metric compared to TopoMLP under the same configuration. 
Notably, UniTopo$\ddagger$ trained with ResNet-50 for 24 epochs even surpasses TopoMLP, which uses Swin-B as the backbone and is trained for 48 epochs, on the $\text{TOP}_{ll}$ metric (30.1\% vs. 28.7\%), demonstrating the strong performance of our model in lane topology reasoning task.

On OpenLane-V2 \textit{subset\_B}, our method also achieves state-of-the-art results across various configurations, significantly surpassing existing methods, as demonstrated in Table~\ref{tab:sota_subsetb}.
Specifically, UniTopo$\ddagger$ trained with ResNet-50 for 24 epochs surpasses $\text{T}^2\text{SG}$ by 8.6\%, while UniTopo$\ddagger$ trained with Swin-B for 48 epochs surpasses TopoMLP by 7.6\%.

Since the lane segment perception task involves the prediction of topological relationships between lane segments, we also compare UniTopo with several state-of-the-art methods on the lane segment perception benchmark. As shown in Table~\ref{tab:laneseg}, UniTopo outperforms LaneSegNet by 6.0\% in terms of the most critical $\text{TOP}_{lsls}$ metric.

\subsection{Ablation Study}

All ablation experiments are conducted on \textit{subset\_A} of the OpenLane-V2 dataset, using ResNet-50 as the backbone, and the models are trained for 24 epochs.

\noindent \textbf{Effectiveness of different modules.} 
To establish a baseline for comparison, we configure TopoNet with parameter settings identical to those of our method, obtaining the results presented in the second row of Table~\ref{tab:module}.
Building upon this baseline, we introduce unified piecewise lane queries and connection queries, as shown in the third row. 
This results in a 10.9\% improvement in the $\text{TOP}_{ll}$ metric.
These substantial gains demonstrate that using connection queries to directly capture topology information from the scene is highly effective in enhancing topology reasoning performance.
Subsequently, we incorporate the Topology-Aware Attention Module~(TAM), which further increases the $\text{TOP}_{ll}$ by 1.7\%. This result indicates that embedding topology information into the lane features can further improve lane topology performance. 
Additionally, we adopt the group-based topology training strategy to accelerate the convergence of the topology head and achieve the results of UniTopo, as shown in the fifth row.
TopoLogic integrates the GeoDist strategy into TopoNet. When using TopoLogic as the baseline, by adding GeoDist to UniTopo, the final results of UniTopo$\ddagger$ are presented in the last row of Table~\ref{tab:module}.

\noindent \textbf{Incorporation of connection queries.}
The Topology-Aware Attention Module (TAM) incorporates lane-to-lane topology information derived from connection queries into lane queries. We further explore alternative implementations for propagating topology information, summarized in Table~\ref{tab:TAM}.
We find that employing a standard self-attention mechanism~(SA) results in degraded model convergence due to the lack of explicit modeling of geometric correlations between piecewise lanes and connected lanes.
Moreover, using geometric distances as edge weights and applying Graph Convolutional Network~(GCN) to model lane-to-lane topology also decreases the performance.
These results suggest that our proposed TAM enables lane queries to more effectively aggregate topological relationship information, thereby enhancing both detection and reasoning performance.

\begin{figure*}[!t]
  \centering
  \includegraphics[width=\linewidth]{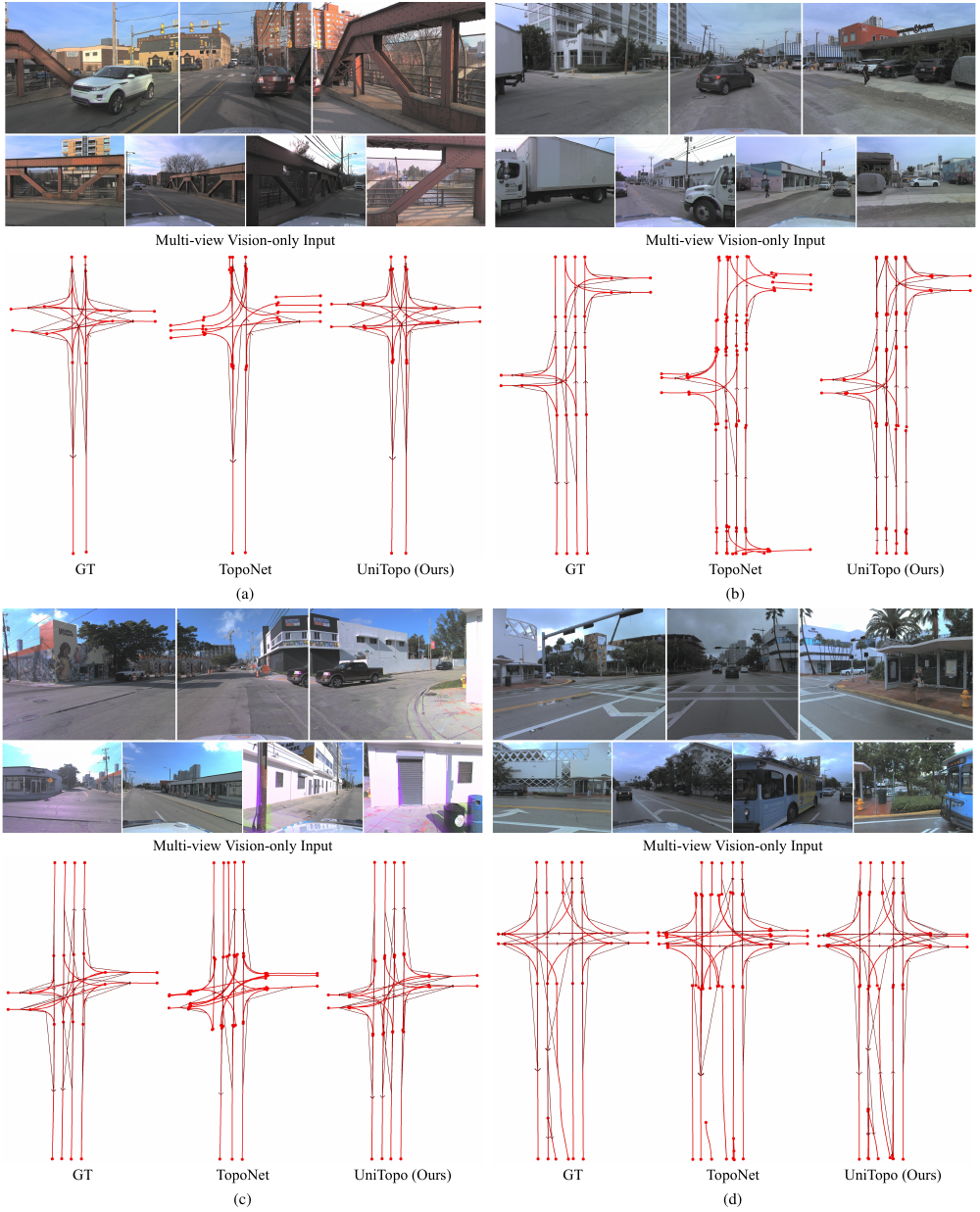}
  \caption{\textbf{Qualitative results across different scenarios} on OpenLane-V2 \textit{subset\_A}. Compared to TopoNet~\cite{TopoNet}, our UniTopo improves lane-to-lane topology accuracy.}
  \label{fig:vis_subset_a}
\end{figure*}

\begin{figure*}[t]
  \centering
  \includegraphics[width=\linewidth]{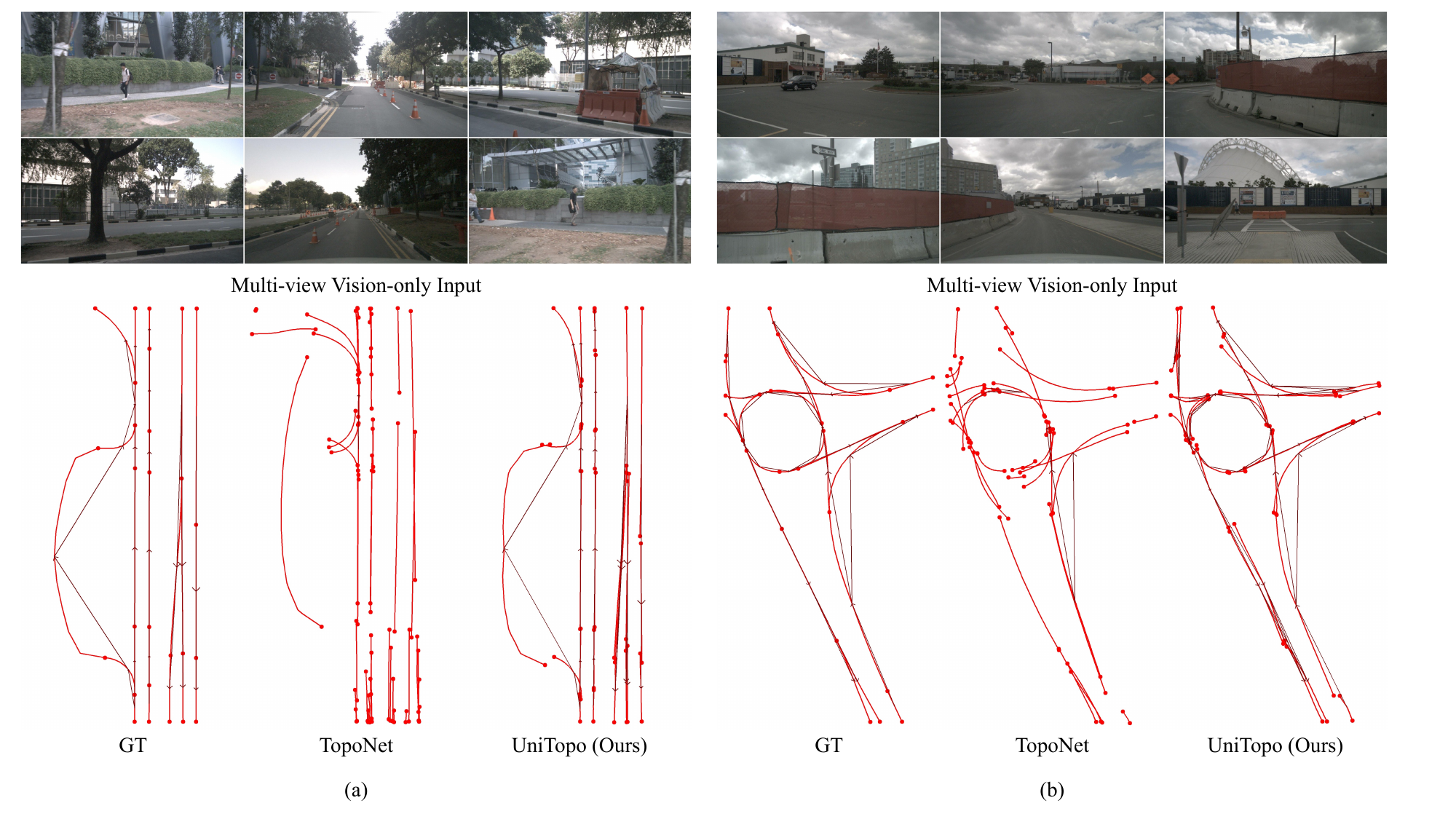}
  \caption{\textbf{Qualitative results of long-tail scenarios} on OpenLane-V2 \textit{subset\_B}. Our UniTopo accurately detects the (a) left-side curved lane and (b) roundabout, and predicts their topological relationships with other lanes.}
  \label{fig:vis_subset_b}
\end{figure*}

\begin{figure}[t]
  \centering
  \includegraphics[width=0.95\linewidth]{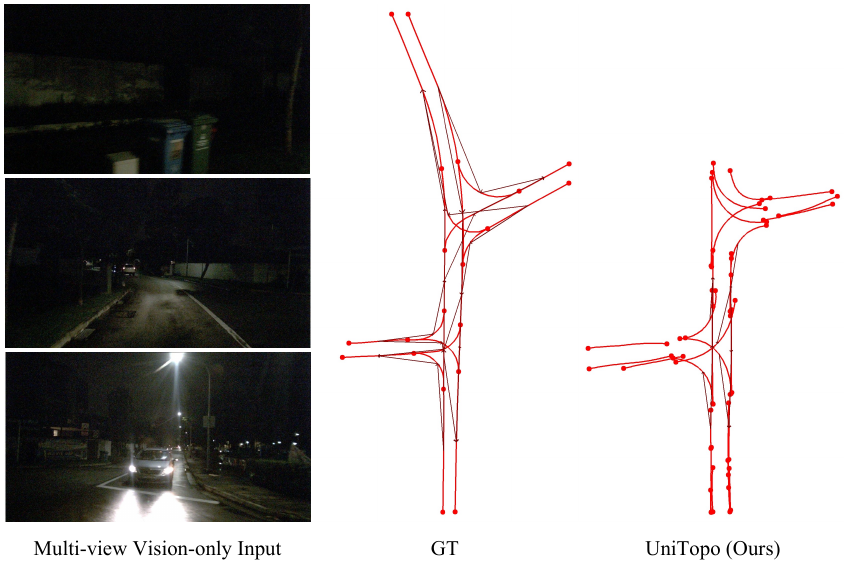}
  \caption{
  \textbf{Failure case} on OpenLane-V2 \textit{subset\_B}. Due to nighttime scenes with low visibility, our method fails to detect distant lanes ahead, along with the corresponding topological relationships. 
  }
  \label{fig:failure_case}
\end{figure}

\noindent \textbf{Number of queries.}
As shown in Table~\ref{tab:lane}, we evaluate the impact of varying the number of lane/connection queries among 200, 300, and 400. In complex scenarios with numerous lanes, using only 200 queries results in a low detection recall rate, which significantly impacts the performance of topology reasoning~(27.0\% vs. 30.1\% on $\text{TOP}_{ll}$). 
However, increasing the number of queries to 400 introduces a higher proportion of negative samples, adversely affecting detection performance~(30.7\% vs. 31.4\% on $\text{DET}_l$). Therefore, we select 300 queries for both lane query and connection query as the default setting in our experiments.

\noindent \textbf{Number of query groups.}
Initially, we design two groups of queries to detect piecewise lanes and connected lanes. To accelerate the convergence of the topology head, we adopt a group-based topology training strategy by introducing additional groups of queries for detection and reasoning during training.
Table~\ref{tab:group} reports the impact of varying the number of groups on model performance. The best results are obtained when the group number is set to 6. As the number of groups increases further, the lane-to-lane topology prediction does not improve, while the performance of traffic element detection gradually declines. 
In particular, when the group number is increased to 8 or 10, the $\text{TOP}_{ll}$ results remain close to those with 6 groups, decreasing slightly by 0.3\% and 0.1\%, respectively. However, the $\text{DET}_{t}$ results drop substantially, by 6.4\% and 7.7\%, which is likely because a larger group number increases the relative weight of lane-related losses, thereby impairing the model's ability to learn traffic element–related information. Based on these observations, we set the group number to 6 as the default setting in our experiments.

\noindent \textbf{Loss weight settings.}
To investigate the impact of different loss items on the final performance, we conduct experiments by individually doubling the weights of detection-related losses (i.e., $\lambda^{l}$ and $\lambda^{t}$), topology-related losses (i.e., $\lambda^{ll}$ and $\lambda^{lt}$), and lane-related losses (i.e., $\lambda^{l}$ and $\lambda^{ll}$). The results are presented in Table~\ref{tab:loss}. The best result is achieved by maintaining the default loss weight settings of TopoNet.

\noindent
\textbf{BEV dependency analysis.}
We analyze the dependency of the proposed framework on BEV representations by jointly varying the BEV encoder depth and the spatial resolution of the constructed BEV feature maps.
Reducing the encoder depth and feature resolution effectively simulates lightweight or degraded BEV representations.
As shown in Table~\ref{tab:bev}, decreasing BEV quality leads to a consistent but gradual performance drop across both detection and topology metrics, rather than a sudden collapse.
These results indicate that the proposed framework is robust to degraded BEV representations and does not critically rely on a specific BEV encoder configuration. 

\noindent
\textbf{Design of attention mask.}
We conduct an ablation study to investigate the design of the attention mask, as shown in Table~\ref{tab:attn}. First, instead of selecting the minimum distance between the piecewise lane and the first and second halves of the connected lane, we use the sum of these two distances to compute the attention mask, which results in 24.9\% $\text{TOP}_{ll}$ and 45.7\% OLS. 
In addition, we remove the parameter sharing of the distance-mapping MLP across decoder layers, resulting in 24.8\% $\text{TOP}_{ll}$ and 45.8\% OLS. 
Both variants lead to performance degradation, thereby confirming the effectiveness of our proposed attention mask design. 

\subsection{Computational Cost Analysis}
We present the computational cost comparison in Table~\ref{tab:flops}.
We measure the runtime of each model with aligned input sizes on a Tesla A100 GPU.
Compared to TopoNet~\cite{TopoNet}, the incorporation of connection queries and the TAM increases the number of parameters and computational cost of our model by 1.74M and 13.98G FLOPs, respectively. However, our model achieves a performance improvement of 18.7\%.
Overall, despite the modest increase in computational cost, our model demonstrates superior performance, confirming its effectiveness and efficiency.
It is worth noting that the reported runtime does not incorporate inference-level optimizations.
The proposed framework is fully compatible with several practical acceleration techniques.
In particular, TensorRT-based graph compilation can be applied to the backbone, BEV encoder, and lane decoder to enable operator fusion and kernel-level optimization.
In addition, INT8 or mixed-precision quantization can be adopted for the convolutional backbone and BEV encoder, which dominate the overall computational cost.
Moreover, the Transformer-style attention modules used in the BEV encoder and topology reasoning stage are well suited for optimized attention kernels such as FlashAttention.
These system-level optimizations are orthogonal to our method and can substantially improve runtime efficiency in practical deployment. 

\subsection{Visualization}
Figure~\ref{fig:vis_subset_a} presents visualization results on OpenLane-V2 \textit{subset\_A}, demonstrating that our model improves lane-to-lane topology accuracy across different scenarios compared to TopoNet.
Besides, we select several long-tail scenarios from OpenLane-V2 \textit{subset\_B}.
As shown in Figure~\ref{fig:vis_subset_b}(a), our model achieves better localization of the curved lane by accurately reasoning about the topology relationships between this lane and adjacent lanes. In the roundabout scenario, our model also accurately predicts the topology relationships among the lanes surrounding the roundabout, as illustrated in Figure~\ref{fig:vis_subset_b}(b).
We further analyze a representative failure case from \textit{subset\_B}, as shown in Fig.~\ref{fig:failure_case}.
This example corresponds to a night scene with severely degraded illumination, which reflects a typical domain shift caused by appearance and lighting changes in real-world deployment.
Under such conditions, reduced visibility and low-contrast appearance primarily affect the image backbone and BEV encoder, leading to weak lane features that propagate to downstream lane detection and topology reasoning.
Importantly, the proposed topology modeling framework operates on structured lane representations and is largely agnostic to visual appearance.
As a result, it is compatible with existing domain-invariant learning techniques, such as category-level adversarial adaptation~\cite{luo2019taking, luo2021category} and pruning-based domain generalization methods~\cite{luo2025kill}, which can be integrated to improve robustness under challenging conditions such as night scenes. 

\section{Conclusion and Discussion}

\subsection{Conclusion}
In this paper, we propose UniTopo, a novel framework for unified modeling of lane and lane topology. 
Inspired by the observation that connections between lanes are clear and observable in original images, we represent lane-to-lane topology as a set of connected lanes. 
By unifying the representation of lanes and their topological relationships, we achieve a unified perception scheme for detecting lane positions and predicting lane topological relationships. 
Experimental results demonstrate that our method achieves 30.1\% and 31.8\% $\text{TOP}_{ll}$ on the two subsets of OpenLane-V2, significantly outperforming previous state-of-the-art methods and demonstrating the strong performance of our method on the lane-to-lane topology prediction task.
\subsection{Limitations and Future Work}
Our research primarily focuses on improving lane-to-lane topology modeling.
While traffic element detection and lane-to-traffic topology prediction are included in the pipeline to comply with existing benchmark settings, we do not introduce new designs or improvements for lane-to-traffic topology modeling.
All traffic element-related components are adopted directly from existing pipelines.
Extending the framework to jointly advance lane-to-traffic topology reasoning remains a promising direction for future work.
We plan to explore broader scene-level modeling, potentially leveraging multimodal large language models to achieve a more comprehensive understanding of complex driving environments. 

\section*{Acknowledgment}
This research is supported in part by the Zhongguancun Academy (Grant No. 20240304), National Key R\&D Program of China (2022ZD0115502), National Natural Science Foundation of China (No. 62461160308, No. 62576024, U23B2010), ``the Fundamental Research Funds for the Central Universities'' (No. 501RCQD2025), ``Pioneer'' and ``Leading Goose'' R\&D Program of Zhejiang (No. 2024C01161), Beijing Natural Science Foundation (QY25227), Ningbo Science and Technology Innovation 2025 Major Project (2025Z034), NSFCRGC Project (N CUHK498/24), and Meituan.

\bibliographystyle{IEEEtran}
\bibliography{main}

\end{document}